\documentclass[letterpaper, 10 pt, conference]{ieeeconf}  % Comment this line out if you need a4paper

\IEEEoverridecommandlockouts                              % This command is only needed if 
\usepackage{times}

\usepackage[nolist]{acronym} % acronyms by the \ac{label} command
\usepackage{amsfonts,siunitx,booktabs} % general useful stuff
\usepackage{graphics} % for pdf, bitmapped graphics files
\usepackage{svg}
\usepackage{epsfig} % for postscript graphics files
\usepackage{mathptmx} % assumes new font selection scheme installed
\usepackage{times} % assumes new font selection scheme installed
\usepackage{amsmath} % assumes amsmath package installed
\usepackage{amssymb}  % assumes amsmath package installed
\usepackage[linesnumbered,ruled]{algorithm2e}
\usepackage{stmaryrd}
\usepackage{soul}
\usepackage{bm}
\usepackage{tikz}
\usepackage{balance}
\usetikzlibrary{positioning, shapes.geometric, arrows}

\usepackage{graphicx}
\usepackage{caption}
\let\labelindent\relax
\usepackage{enumitem}
\usepackage{svg}
\usepackage{float}
\usepackage[normalem]{ulem}
\usepackage{hyperref}
\usepackage{algpseudocode}

\usepackage{scalerel}
\usepackage{tikz,standalone,pgfplots}
\usepackage[framemethod=TikZ]{mdframed}
\usetikzlibrary{shadows}
\usepackage{lipsum} % for generating dummy text
\usepackage{overpic}

\DeclareMathAlphabet{\pazocal}{OMS}{zplm}{m}{n}

\makeatletter
\let\@oldmaketitle\@maketitle % Store \@maketitle
\renewcommand{\@maketitle}{\@oldmaketitle % Update \@maketitle to insert...
  \setcounter{figure}{0}
  \vspace{1em}
  \centering    
  \includegraphics[width=0.9\linewidth]{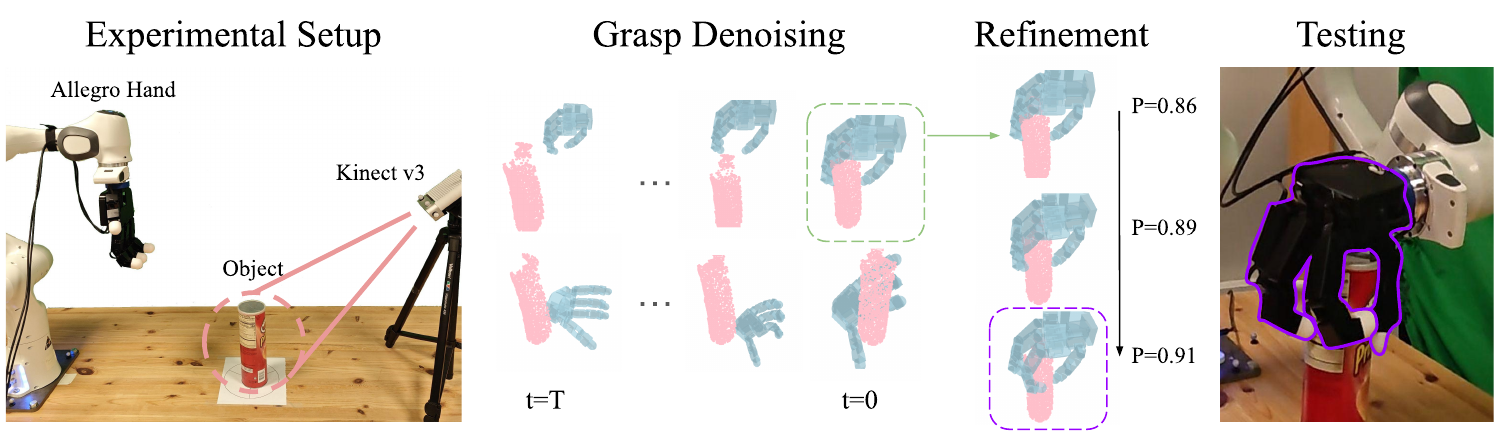}
  \captionof{figure}{\textbf{\methodname{} pipeline}. Given a partial point cloud captured by a Kinect v3, \sampler{} generated a set of high-quality grasps by gradually removing noise from randomly sampled grasps. Subsequently, the grasps were refined and ranked by the score $\prob{}$ from \evaluator{}. The grasp with the highest score was finally selected and executed on the real robot.}
  \label{fig:pull_figure}
}
\makeatother

\title{\LARGE \bf
DexDiffuser: Generating Dexterous Grasps with Diffusion Models
}

\author{Zehang Weng$^{*}$, Haofei Lu$^{*,\dag}$, Danica Kragic, and Jens Lundell% <-this % stops a space
\thanks{$^*$Authors with equal contribution. $^\dag$ Corresponding author.}
\thanks{All authors are with the division of Robotics, Perception, and Learning (RPL) at KTH, Stockholm, Sweden. 
        {\tt\small \{zehang,haofeil,jelundel,dani\}@kth.se}}%
}

\begin{document}
\newcommand{\jensrmk}[1]{{\color{blue} {[\bf j: #1]}}}
\newcommand{\jens}[1]{{\color{blue}#1}}
\newcommand{\zehangrmk}[1]{{\color{green} {[\bf f: #1]}}}
\newcommand{\zehang}[1]{{\color{blue}#1}}
\newcommand{\haofeirmk}[1]{{\color{cyan} {[\bf v: #1]}}}
\newcommand{\haofei}[1]{{\color{cyan}#1}}

\newcommand{\equationref}[1]{\hyperref[#1]{Eq.~\ref*{#1}}}
\newcommand{\figref}[1]{\hyperref[#1]{Fig.~\ref*{#1}}}
\newcommand{\tabref}[1]{\hyperref[#1]{Table~\ref*{#1}}}
\newcommand{\secref}[1]{\hyperref[#1]{Section~\ref*{#1}}}
\newcommand{\algoref}[1]{\hyperref[#1]{Algorithm~\ref*{#1}}}
\newcommand{\figsref}[2]{Figures~\ref{#1}-\ref{#2}}
\newcommand{\subfigref}[1]{(\subref{#1})}

\newcommand{\norm}[1]{\left\lVert#1\right\rVert}
\newcommand{\Var}{\mathrm{Var}}
\newcommand{\ra}[1]{\renewcommand{\arraystretch}{#1}}
\newcommand{\tbs}[1]{\renewcommand{\tabcolsep}{#1pt}}
\newcommand{\abs}[1]{\left\lvert#1\right\rvert}

\newcommand{\matr}[1]{\mathbf{#1}}
\newcommand{\argmax}{\operatornamewithlimits{argmax}}
\newcommand{\argmin}{\operatornamewithlimits{argmin}}
\newcommand*{\prob}{\mathsf{P}}
\newcommand{\de}[1]{\operatorname{d}\!#1}
\newcommand{\etal}[1]{#1 et al.}

\def\sampler{DexSampler}
\def\evaluator{DexEvaluator}
\def\ffhnet{FFHNet}
\def\unidexgrasp{UniDexGrasp}
\def\datasetname{CONG}
\def\datasetacronomy{Acronym}
\def\franka{Franka Emika Panda}
\def\kinect{Kinect 3.0}
\def\graspnet{GraspNet}
\def\pointnet{PointNet++}
\def\graspnetta{GraspNet \ac{tai}}
\def\gonet{GoNet}
\def\gpd{GPD}
\def\edgegrasp{EdgeGraspNet}
\def\equiv{$\mathbb{R}^3\times \text{SO(2)-equivariant}$}
\def\methodname{DexDiffuser}

\def\isaacgym{Isaac Gym}
\def\bestcolor{(best viewed in color)}
\def\sota{state-of-the-art}
\def\ie{, \textit{i.e.}, }
\def\eg{, \textit{e.g.}, }
\def\pc{point cloud}
\def\pcs{point clouds}
\def\epst{\multicolumn{1}{c}{$\epsilon$}}
\def\vt{\multicolumn{1}{c}{$v$}}
\def\nat{\multicolumn{1}{c}{--}}
\newcommand{\blue}[1]{\textcolor{black}{#1}}
\def\samplershort{DexS}
\def\DexGNDataset{DexGN-dataset}
\def\DexGNSampler{DexGN-sampler}
\def\DexGNDatasetShort{DexGN-D}
\def\DexGNSamplerShort{DexGN-S}

\newacro{dnn}[DNN]{Deep Neural Network}
\newacro{fcn}[FCN]{Fully Convolutional Network}
\newacro{sdf}[SDF]{Signed Distance Function}
\newacro{cnn}[CNN]{Convolutional Neural Network}
\newacro{gnn}[GNN]{Graph Neural Network}
\newacro{dl}[DL]{Deep Learning}
\newacro{ml}[ML]{Machine Learning}
\newacro{mc}[MC]{Monte Carlo}
\newacro{mlp}[MLP]{Multi-Layer Perceptron}
\newacro{dof}[DoF]{Degrees of Freedom}
\newacro{vae}[VAE]{Variational Autoencoder}
\newacro{cvae}[CVAE]{Conditional Variational Autoencoder}
\newacro{fps}[FPS]{Farthest Point Sampling}
\newacro{tai}[TaI]{Target as Input}
\newacro{pca}[PCA]{Principal Component Analysis}
\newacro{pc}[PC]{Principal Component}
\newacro{auc}[AUC]{Area Under the Curve}
\newacro{elbo}[ELBO]{Evidence Lower Bound}
\newacro{egd}[EGD]{Evaluator-Guided Diffusion}
\newacro{esr}[ESR]{Evaluator-based Sampling Refinement}
\newacro{rl}[RL]{Reinforcement Learning}
\newacro{bps}[BPS]{Basis Point Set}
\newacro{mh}[MH]{Metropolis-Hasting}
\newacro{fc}[FC]{Fully Connected}

\maketitle

\thispagestyle{empty}
\pagestyle{empty}

%%%%%%%%%%%%%%%%%%%%%%%%%%%%%%%%%%%%%%%%%%%%%%%%%%%%%%%%%%%%%%%%%%%%%%%%%%%%%%%%
\begin{abstract}

We introduce \methodname{}, a novel dexterous grasping method that generates, evaluates, and refines grasps on partial object point clouds. \methodname{} includes the conditional diffusion-based grasp sampler \sampler{} and the dexterous grasp evaluator \evaluator{}. \sampler{} generates high-quality grasps conditioned on object point clouds by iterative denoising of randomly sampled grasps. We also introduce two grasp refinement strategies: \acl{egd} and \acl{esr}. The experiment results demonstrate that \methodname{} consistently outperforms the state-of-the-art multi-finger grasp generation method \ffhnet{} with an, on average, 9.12\% and 19.44\% higher grasp success rate in simulation and real robot experiments, respectively. Supplementary materials are available at \url{https://yulihn.github.io/DexDiffuser_page/}.
\end{abstract}

\section{Introduction}

Despite years of research on data-driven grasping \cite{newbury2023deep}, generating dexterous grasps for picking unknown objects, similar to the example shown in \figref{fig:pull_figure}, remains challenging \cite{mayer2022ffhnet}. One of the main challenges in dexterous grasping is identifying successful grasps within a high-dimensional search space \cite{Ciocarlie2007DexterousGV}. 

In this work, we propose the new data-driven dexterous grasping method \methodname{} to address the dimensionality issue. \methodname{} includes \sampler{}, a new conditional diffusion-based dexterous grasp sampler, and \evaluator{}, a new dexterous grasp evaluator. \sampler{} generates high-quality grasps conditioned on object point clouds by iteratively denoising randomly sampled grasps. We trained the models on 1.7 million successful and unsuccessful Allegro Hand grasp generated using DexGraspNet \cite{dexgraspnet} across 5378 objects. In addition to \methodname{}, we also propose two grasp refinement strategies: \acf{egd}, which is only applicable for diffusion models, and \acf{esr}, which applies to all dexterous grasp sampling methods.

We experimentally assess \methodname{}'s capability to sample, evaluate, and refine 16-\ac{dof} Allegro Hand grasps on three visually and geometrically distinctive object datasets in simulation and on one dataset in the real world. \methodname{} is benchmarked against the \sota{} contactmap-based method \unidexgrasp{} \cite{Xu_2023_CVPR} in simulation and the multi-finger grasp generation method \ffhnet{} \cite{mayer2022ffhnet} in simulation and the real world. The experimental results indicate that our best method achieves a grasp success rate of 98.77\% in simulation and 68.89\% in the real world, which is 11.63\% and 20.00\% higher than the respectively best \ffhnet{} model.  

Our contribution can be summarized as follows:
\begin{itemize}
    \item \sampler{} the novel diffusion-based dexterous grasp sampler  (\secref{sec:generator}), and \evaluator{} the dexterous grasp evaluator (\secref{sec:evaluator}).
    \item The two refinement strategies \ac{egd} and \ac{esr} for improving dexterous grasps (\secref{sec:improvement}).
    %\item A dataset of 1.7 million successful and unsuccessful Allegro Hand grasp across 5378 objects (\secref{sec:dataset}).
    \item A comprehensive experimental evaluation demonstrating the efficacy and real-world applicability of \methodname{} (\secref{sec:exp}).
\end{itemize}

\section{Related Work}
\label{sec:related_work}

This work spans two topics: data-driven dexterous grasping and diffusion models. For brevity, we limit the related works on diffusion models to robotics and, for other areas, refer to the comprehensive survey by \etal{Yang} \cite{yang2023diffusion}.

\subsection{Data-Driven Dexterous Grasping}

Data-driven dexterous grasp sampling methods primarily fall into three categories: 1) those that generate a contact map on the object's surface \cite{Xu_2023_CVPR, li2022gendexgrasp, lu2023ugg, wu2022learning}, 2) those that rely on shape completion \cite{ottenhaus2019visuo, lundell2021multi, jensdgcc, van2020learning, lu2020multi, lu2020multifingered,liu2019generating,wei2022dvgg},  or 3) those that train a grasping policy based on \ac{rl} \cite{qin2023dexpoint,mandikal2021graff} or human demonstrations \cite{wan2023unidexgrasp}. The main limitation of the methods in 1) is the assumption of complete object observation, which hinders real-world applications where complete visibility cannot be guaranteed. Methods in 2) are limited because the generated grasp quality is closely intertwined with the completion quality while collecting expensive demonstrations, and the sim-to-real gap limits the methods in 3).

Only a few methods \cite{popovic2011grasping,choi2018learning, mayer2022ffhnet} do not adhere to any of the above categories as they generate grasp poses directly on incomplete object observations without shape completion nor training a grasping policy. Our early work \cite{popovic2011grasping}, focuses on building hierarchical representations of objects on which grasps are sampled. To make learning more tractable, \etal{Choi} \cite{choi2018learning}  considered one gripper configuration and discretized grasp poses into 6 approach directions and 4 orientations. A discriminative network is trained to score the 24 pre-selected grasp poses for the highest grasp success probability.  \etal{Qian} \cite{mayer2022ffhnet} proposed \ffhnet{}: a \ac{vae}-based grasp sampler that generates 15-\ac{dof} gripper configurations and grasp poses in all of SE(3). Albeit \ffhnet{} reached a 91\% grasp success rate in simulation, it was never evaluated on real hardware, leaving the question of real-world performance unanswered. \methodname{} also produces complete gripper configurations and SE(3) grasp poses but by using diffusion models. Also, we experimentally validate \methodname{} and \ffhnet{} on real robotic hardware. 

%Our experimental evaluations in simulation and the real world show significant improvement of \methodname{} over \ffhnet{}.  

\subsection{Diffusion Models in Robotics}

Despite beind rather recent, diffusion models have already found widespread applicability in motion planning \cite{carvalho2023motion}, navigation \cite{ryu2023diffusion}, manipulation \cite{chi2023diffusionpolicy,structdiffusion2023,simeonov2023rpdiff,chen2023playfusion}, human-robot interaction \cite{ng2023diffusion} and grasping \cite{urain2023se}. Of these, the most related to ours is the grasping work by \etal{Urain} \cite{urain2023se}, where a diffusion model was trained to generate SE(3) parallel-jaw grasp poses. We propose a diffusion model for generating \textit{dexterous} grasps, meaning that \textit{both} SE(3) grasp poses and high dimensional gripper configurations are generated.

Prior works on diffusion models in robotics have also trained discriminators to evaluate the diffusion-generated samples \cite{structdiffusion2023,simeonov2023rpdiff}. In \cite{structdiffusion2023}, the discriminator scores how realistic point cloud-generated scenes are, while the discriminator in \cite{simeonov2023rpdiff} scores how good a generated SE(3) pose is for manipulating an object. Similarly, we use a discriminator to evaluate a generated grasp's success probability in picking an object. 
\section{Problem Statement}
\label{sec:problem_statement}

We consider the problem of sampling and evaluating dexterous grasps $\matr{G}$ on partially observed point clouds $\matr{O}\in \mathbb{R}^{\text{N}\times 3}$, where $N$ is the number of points, for the task of successfully picking up objects. In this work, a grasp $ \matr{g} \in \matr{G} = [\matr{p}, \matr{r}, \matr{q}] \in \mathbb{R}^{9+k}$ is represented by a 3-D position  $\matr{p} \in \mathbb{R}^3$, a continouous 6-D rotation vector $ \matr{r} \in \mathbb{R}^6$ \cite{zhou2019continuity}, and a gripper joint configuration $\matr{q} \in \mathbb{R}^k$, where $k$ is the number of controllable joints. Henceforth, $\matr{p}$ and $\matr{r}$ are together referred to as the grasp pose and $\matr{q}$ as the joint angles. We assume all objects are reachable and singulated, and the object point cloud is segmented from the scene.

We frame the problem of sampling grasps as learning a distribution $\prob{(\matr{g} \in \matr{G} | \matr{O})}$ and of evaluating grasps as learning a discriminator $\prob(S=1|\matr{g}, \matr{O})$, where $S$ is a binary variable representing grasp success ($S$=1) and failure ($S$=0). Henceforth, $\prob{(S=1 | \matr{g} \in \matr{G}, \matr{O})}$ is referred to as the grasp evaluator and  $\prob{(\matr{g} \in \matr{G} | \matr{O})}$ as the grasp generator. We approximate each of these distributions with a separate \ac{dnn} $\mathcal{D}_{\boldsymbol{\theta}}(\matr{g} \in \matr{G}|\matr{O})\approx \prob{(\matr{g} \in \matr{G}|\matr{O})}$ and $\mathcal{D}_{\boldsymbol{\psi}}(S=1 | \matr{g} \in \matr{G}, \matr{O}) \approx \prob{(S=1 | \matr{g} \in \matr{G}, \matr{O})}$, with learnable parameters $\boldsymbol{\theta}$ and $\boldsymbol{\psi}$. The objective then becomes learning $\boldsymbol{\theta}$ and $\boldsymbol{\psi}$ from data.
\section{Method}

We now introduce our framework \methodname{} to learn the parameters $\boldsymbol{\theta}$ and $\boldsymbol{\psi}$. \methodname{} consists of two parts: \sampler{} and \evaluator{}. We also propose two grasp refinement strategies for improving the grasp success probability.

\subsection{The Basis Point Set Representation}\label{sec:bps}

\methodname{} generates and samples grasps directly on partial object point clouds $\matr{O}$. In this work, $\matr{O}$ is encoded into $\matr{f}_\matr{O}$ using \ac{bps} \cite{prokudin2019efficient} as this encoding has shown to work well in prior dexterous grasping work \cite{mayer2022ffhnet}. The \ac{bps} encoding represents the shortest distance between each point in a fixed set of basis points and all points in $\matr{O}$. The benefit of \ac{bps} is that it captures the shape and spatial properties of the object and is of fixed length, which allows for the use of computationally efficient \ac{fc} layers to further process $\matr{f_{O}}$. 

\subsection{Diffusion-based Grasp Sampler}\label{sec:generator}
% \vspace{0.01cm} % Adjust the space before the image
\begin{figure*}[ht]
    \centering
\includegraphics[width=0.9\linewidth]{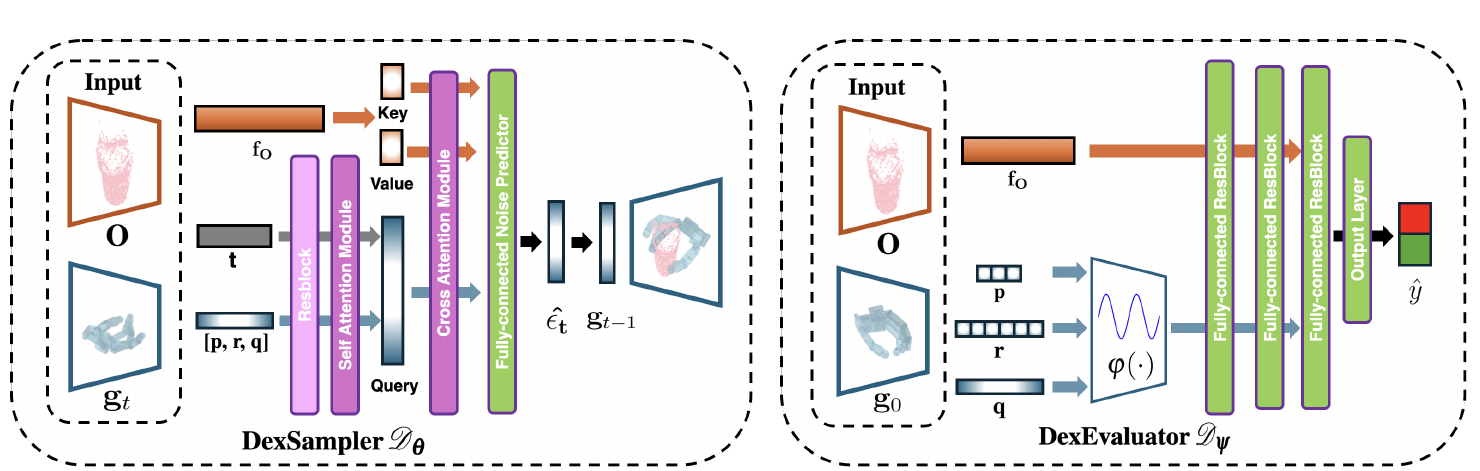}
\caption{\textbf{Model architecture}. The left image shows the \sampler{}. Its input $\matr{f_{O}}$ is processed into a key-value pair while $\matr{g}_t$ and $t$ are processed into a query using a self-attention block. The key-value-query triplet is then embedded using a cross-attention block to compute $\hat{\matr{\epsilon}}_t$. The right image shows the \evaluator{} that predicts grasp success probability given the same $\matr{f_{O}}$ as the \sampler{} and $\matr{g}_0$ produced by the \sampler{}.}
\label{fig:dexdiffsuser_architecture}
% \vspace{-10px}
\end{figure*}

\sampler{} is a classifier-free, conditional diffusion model that generates $\matr{G}$ conditioned on $\matr{f_{O}}$. 
It consists of a forward process that gradually turns data into Gaussian noise and a learnable inverse diffusion process that recovers the data from noise given the condition \cite{ho2022classifier}.
%using a noise predictor $\matr{\epsilon_{\matr{\theta}}}$ that produce the noisy component of $\matr{g}{_{t}}$. 

In our case, the forward process for incrementally diffusing a successful grasp $\matr{g}_0=[\matr{r}_0, \matr{p}_0, \matr{q}_0]$ over $T$ timesteps is
\begin{align}
\label{eq:forward_diff}
    q(\matr{g}_{1:T}|\matr{g}_0) &= \Pi_{t=1}^{T} q(\matr{g}{_{t}}|\matr{g}_{t-1}), \\
    q(\matr{g}{_{t}}|\matr{g}{_{t-1}})&=\mathcal{N}(\matr{g}{_{t}};\sqrt{1-\beta_{t}}\matr{g}{_{t-1}}, \beta_{t} \matr{I}),
\end{align}
where $\beta_{t}$ is the scheduled noise variance at time step $t$. To recover the original $\matr{g}_0$ from $\matr{g}_T$, \sampler{} learns the reverse diffusion steps by estimating the Gaussian noise added at each $t$ and then removes it as
\begin{align}
\label{eq:inv_diff}
    p_{\boldsymbol{\theta}}(\matr{g}{_{0:T}}|\matr{f_{O}}) &= p(\matr{g}{_{T}})\Pi_{t=1}^{T} p_{\boldsymbol{\theta}}(\matr{g}{_{t-1}}|\matr{g}{_{t}}, \matr{f_{O}}), \\
    p_{\boldsymbol{\theta}}(\matr{g}{_{t-1}}|\matr{g}{_{t}}, \matr{f_{O}}) &= \mathcal{N}(\matr{g}{_{t-1}}; \hat{\mu}_{\boldsymbol{\theta}}(\matr{g}_t,t,\matr{f}{_{O}}), \hat{\Sigma}_{{\boldsymbol{\theta}}}(\matr{g}_t,t,\matr{f}{_{O}})),
\end{align}
where $\hat{\mu}_{\boldsymbol{\theta}}(\matr{g}_t,t,\matr{f}{_{O}})$ and $\hat{\Sigma}_{{\boldsymbol{\theta}}}(\matr{g}_t,t,\matr{f}{_{O}})$ is the learnable mean and variance of a Gaussian distribution. The parameters $\boldsymbol{\theta}$ are learned by minimizing the loss function
\begin{align}
\label{eq:noise_pred}
    L_{\epsilon} = ||\hat{\matr{\epsilon}}{_{t}} - \matr{\epsilon}{_{t}}||^2,
\end{align}
%where $\hat{\matr{\epsilon}}{_{t}}\sim \mathcal{N}(\matr{g}{_{t-1}}; \hat{\mu}_{\boldsymbol{\theta}}(\matr{g}_t,t,\matr{f}{_{O}}), \hat{\Sigma}_{{\boldsymbol{\theta}}}(\matr{g}_t,t,\matr{f}{_{O}}))$ and $\matr{\epsilon}_{t}$ is the ground-truth noise.
where $\hat{\matr{\epsilon}}{_{t}}=\epsilon_{\theta}(\matr{g}_t,t,\matr{f}{_{O}})$ and $\matr{\epsilon}_{t}$ is the ground-truth noise.

The architecture of \sampler{} is depicted in \figref{fig:dexdiffsuser_architecture}, following the practical design of conditional diffusion model \cite{ ho2022cascaded,rombach2022high,saharia2022photorealistic,huang2023diffusion}. 
It takes three inputs: $\matr{f_{O}}$, $\matr{g}_{t}$, and $t$ and outputs $\matr{\hat{\epsilon}}_{t}$ for grasp denoising.

\subsection{Grasp Evaluator}\label{sec:evaluator}
Although \methodname{} is trained to generate successful grasps, it can still produce unsuccessful ones. Thus, it would be useful if we could only execute grasps with a high probability of succeeding. To this end, we propose \evaluator{} $\mathcal{D}_{\boldsymbol{\psi}}(S=1 | \matr{g} \in \matr{G}, \matr{f_{O}})$ to predict the grasp success probability. The architecture of \evaluator{} is shown in \figref{fig:dexdiffsuser_architecture} and is inspired by the evaluator in \cite{mayer2022ffhnet}. It takes two inputs: $\matr{g}$ and $\matr{f_{O}}$ and outputs $S$. The main difference to \cite{mayer2022ffhnet} is that we introduce frequency encoding \cite{mildenhall2020nerf} that maps $\matr{g}$ into a higher dimensional space before passing it to the evaluator. The map takes the following form  
\begin{align}
\label{eq:eval_pos_enc}
    \varphi( \matr{x} ) = &\lbrack\sin(2^0\pi \matr{x}),~\cos(2^0\pi \matr{x}),~\ldots, \nonumber \\
     &\sin(2^{\text{F}-1}\pi \matr{x}),~\cos(2^{\text{F}-1}\pi \matr{x})\rbrack,
\end{align}
where F specifies the number of sinusoidal functions. $\varphi(\cdot)$ is then separately applied on $\matr{p}$, $\matr{r}$, and $\matr{q}$. Introducing frequency mapping into the evaluator allows it to model better how small changes in $\matr{g}$ can lead to large changes in $S$ \cite{mousavian2019graspnet}, which requires the model to be sensitive to high-frequency details \cite{mildenhall2020nerf}. %Introducing the position encodings into the evaluator addresses the problem that of modelling the small changes in $\matr{g}$ can lead to large changes in $\prob{(S | \matr{g} \in \matr{G}, \matr{O})}$ \cite{mousavian2019graspnet}. This  which is similar to  .

The parameters $\boldsymbol{\psi}$ of the \evaluator{} is optimized using the binary cross-entropy loss

\begin{align}
\label{eq:eval_loss}
    L(y, \hat{y}) = -y \log(\hat{y}) + (1 - y) \log(1 - \hat{y}),
\end{align}
where $y$ and $\hat{y}$ denote the ground truth and predicted grasp labels, respectively. In \secref{sec:dataset}, we provide details on the dataset used to train \evaluator{}.

In addition to evaluating grasps, the \evaluator{} can also be used to guide the inverse diffusion process towards more successful grasps or refine already sampled grasps. This we discuss next.

\subsection{Grasp Refinement}\label{sec:improvement}

The idea behind grasp refinement is to increase the probability of grasp success. In this work, we propose two grasp refinement strategies: 
\ac{egd}, which uses the evaluator to guide the inverse diffusion process, and \ac{esr}, which uses the evaluator to refine already sampled grasps.

\subsubsection{\ac{egd}}
\ac{egd} is a form of classifier-guided diffusion. Formally, during the inverse diffusion process, a noisy grasp $\matr{g}{_{t}}$ is iteratively denoised as
\begin{align}
\label{eq:egd}
    p_{\theta}(\matr{g}{_{t-1}}|\matr{g}{_{t}}, \matr{f_{O}}) = & \mathcal{N}\left(\matr{g}{_{t-1}};\mu_{\theta}(\matr{g}{_{t}},t,\matr{f}{_{O}}) + \lambda \cdot \matr{g}{}{_\text{guide}}, \right. \nonumber \\
    &\left. \vphantom{\mathcal{N}} \Sigma_{\theta}(\matr{g}{_{t}},t,\matr{f_{O}}) \right), \\
    \matr{g}{_\text{guide}} = &\log(T-t + 1)\nabla_{\matr{g}{_t}} \log \mathcal{D}_{\phi}(\hat{y}=1|\matr{g}{_t}, \matr{f_{O}}) ,
\end{align}
where $\matr{g}{_\text{guide}}$ is the gradient signal from \evaluator{} modulated by $\log(T-t + 1)$. In each step of the inverse diffusion process, $\matr{g}{_\text{guide}}$ is used to adjust $\mu_{\theta}$ towards more successful grasp. The magnitude of the adjustment is controlled by the parameter $\lambda$. The time-dependent modulation term $\log(T-t + 1)$ ensures that the guidance is more pronounced in the earlier stages of denoising when the uncertainty in the grasp is higher.

\subsubsection{\ac{esr}}
Compared to \ac{egd}, \ac{esr} locally refines already sampled grasps. Prior research on parallel-jaw grasp sampling 
\cite{mousavian2019graspnet} has demonstrated that many sampled grasps with low evaluator scores often lie spatially close to higher-scoring ones, suggesting that local grasp refinement can improve the grasp success probability. Building upon this insight, we extend the concept of local refinement for parallel-jaw grasps \cite{mousavian2019graspnet,murali20206} to dexterous grasps, which requires us to refine $\matr{p}$, $\matr{r}$ \textit{and} $\matr{q}$. 

Mathematically, local grasp refinement boils down to finding a $\Delta \matr{g}$ so that $\prob{(S=1 | \matr{g} + \Delta \matr{g},~\matr{f_{O}})}>\prob{(S=1 | \matr{g},~\matr{f_{O}})}$. To find $\Delta \matr{g}$, we use the \ac{mh} sampling algorithm as it has been shown to perform well for refining parallel-jaw grasps \cite{murali20206}. To apply \ac{mh}, we first sample  $\Delta \matr{g}$ from a known probability distribution and then calculate the acceptance ratio $\alpha=\frac{\mathcal{D}_{\boldsymbol{\psi}}(S=1 | \matr{g}+\Delta \matr{g}, \matr{f_{O}})}{\mathcal{D}_{\boldsymbol{\psi}}(S=1 | \matr{g}, \matr{f_{O}})}$. If $\alpha\geq u$, where $u\sim\mathcal{U}[0,1]$, then $\matr{g}=\matr{g}+\Delta \matr{g}$ otherwise $\matr{g}$ is not updated. We refer to this process as \ac{esr}-1, as it performs refinement simultaneously for every element of $\matr{g}$.

Because of the higher dimensional search space, local refinement is more challenging for dexterous than parallel-jaw grasps. Therefore, to make the refinement process more tractable, we introduce a two-stage \ac{esr} (\ac{esr}-2) that first refines $\matr{p}$ and $\matr{r}$ jointly and only afterward refines $\matr{q}$. The idea behind \ac{esr}-2 is first to refine the global grasp parameters $\matr{p}$ and $\matr{r}$ and only when these are good, refine the local parameters $\matr{q}$.

\subsection{Implementation Details}

For training \sampler{}, the hyperparameter $\beta_t$ is controlled using a linear scheduler that moves from \num{1e-4} to \num{1e-2} over a total of 100 timesteps. The learning rate is initially set to \num{1e-4} and controlled by an exponential rate scheduler with $\gamma=0.9$. \sampler{} is trained with a batch size of 16384 for 200 epochs. The learning rate for the \evaluator{} is also set to \num{1e-4} but controlled with a plateau scheduler. The \evaluator{} is trained with a batch size of 25600 object-grasp pairs for 20 epochs. 

For calculating the \ac{bps} we randomly sample one set of 4096 basis points. This set is then used to calculate the \ac{bps} for every $\matr{O}$ in the dataset.

\section{Dataset}\label{sec:dataset}

We used DexGraspNet \cite{dexgraspnet} to generate training grasps and Isaacgym to render 50 $\matr{O}$ from different camera views for each object. We collected over 0.7M successful and 0.8M unsuccessful Allegro Hand grasps on 5378 objects of various geometries and scales. Only successful grasps were used to train \sampler{} while both successful and unsuccessful grasps were used to train \evaluator{}. 

To enhance the dataset with out-of-distribution negative samples, we generated more grasp candidates by jittering the successful grasps in the original dataset. Specifically, a grasp perturbation $\Delta \matr{g}_k^\text{{pert}}$ was added for every successful object-grasp pair $(\matr{O}_k, \matr{g}_k)$. We tested and labeled all the perturbed grasps in Isaacgym and only kept the unsuccessful ones.

The final dataset consisted of 1.7 million grasps, where approximately 40.52\% were successful and 59.48\% were unsuccessful. We split this dataset into a training set of 4303 objects and a test set of 1,075. We further add 49 objects from the EGAD! dataset \cite{morrison2020egad}, 16 objects from the MultiDex dataset \cite{li2022gendexgrasp}, and 12 objects from the KIT dataset \cite{kasper2012kit} to the the test dataset. 

\section{Experimental evaluation}\label{sec:exp}

We experimentally evaluated \methodname{} in both simulation and the real world. %, comparing it to the state-of-the-art grasp sampler and evaluator \ffhnet{} \cite{mayer2022ffhnet}. 
The specific questions we wanted to address with the experiments were:
\begin{enumerate}
    \item How well can the \evaluator{} predict grasp success?
    % \item What effect does the \ac{bps}-encoding have on generating successful grasps?
    \item How well can \sampler{} generate successful and diverse grasps?
    \item To what extent do \ac{esr} and \ac{egd} improve the grasp quality?
    \item How well can \methodname{} generalize to real-world object grasping?
\end{enumerate}

\subsection{Simulation Experiments}
All simulation experiments were conducted in Isaacgym \cite{makoviychuk2021isaac} following the setup from \cite{li2022gendexgrasp}\footnote{It is worth pointing out that the criteria is less strict than \cite{dexgraspnet}.}. Specifically, test objects were initialized as free-floating objects, and a virtual 16-\ac{dof} Allegro Hand was used for grasping each of them. For each of the 1152 test objects, 10 point clouds were rendered from randomly sampled camera poses. Then, all methods generated 20 grasps per point cloud, which amounted to 230,400 test grasps in total. Once an object was grasped, the grasp stability was assessed by applying external forces along the six directions in space for 50 timesteps. A grasp succeeded if the object remained in the hand throughout the stability test.

We also trained a \sampler{} with the PointNet++ point cloud encoding \cite{qi2017pointnet++} to study how different encodings affect grasp performance. We call that network \methodname{}-PN2 while the \ac{bps} network is referred to as \methodname{}-\ac{bps}.

\begin{table}[t]
    \centering
    \begin{adjustbox}{max width=\linewidth}
         \begin{tabular}{lcccc}
            \toprule
            \textbf{Evaluator} & \textbf{Recall Pos. (\%)$\uparrow$} & \textbf{Recall Neg. (\%)$\uparrow$} & \textbf{Total Acc. (\%)$\uparrow$} \\
            \midrule
            No Freq. Enc. & 56.47  & 90.88 & 73.68 \\
            $(10,0,0)$ & 77.28 & 83.42 & 80.35 \\
            $(0,10,0)$ & 54.19 & 91.35 & 72.77 \\
            $(0,0,10)$ & 77.13 & 78.22 & 77.65 \\
            $(5,5,5)$ & 55.49 & \textbf{91.78} & 73.64 \\
            $(10,10,10)$ & 75.92 & 78.01 & 76.97 \\
            $(10,4,4)$ & 55.32 & 91.54 & 73.43 \\
            $(20,4,0)$ & 74.53 & 83.68 & 79.10 \\
            $(10,4,0)$ & \textbf{80.85} & 80.97 & \textbf{80.91}  \\
            \bottomrule
        \end{tabular}
    \end{adjustbox}
    \caption{\textbf{Grasp success prediction}. $(f_1, f_2, f_3)$ represent $F$ in \eqref{eq:eval_pos_enc} for $\matr{p}$, $\matr{r}$, and $\matr{q}$, respectively. $\uparrow$: The higher, the better.}
    \label{tab:sim_exp_evaluator}
\end{table}

\subsubsection{Grasp Evaluation}

To answer the first question, we assessed the \evaluator{}'s grasp prediction accuracy over different numbers of position encodings. The results are presented in \tabref{tab:sim_exp_evaluator}. The findings indicate that the \evaluator{} configured with frequencies $(10,4,0)$ achieves the best performance and maintains a balanced accuracy between positive and negative predictions. \evaluator{} with few or no frequencies is biased towards negative predictions, while increasing the number of frequencies from 10 to 20 reduces performance. Thus, we choose the $(10,4,0)$ \evaluator{} in all subsequent experiments.

\subsubsection{Grasp Sampling}

\begin{table}[t]
    \centering
    \begin{adjustbox}{max width=\linewidth}
         \begin{tabular}{lccccc}
            \toprule
            \textbf{Method} & \multicolumn{4}{c}{\textbf{Dataset}} \\
            \cmidrule(l){2-5}
            & \textbf{MultiDex (\%)$\uparrow$} & \textbf{EGAD! (\%)$\uparrow$} & \textbf{\DexGNDatasetShort{} (\%)$\uparrow$} & \textbf{KIT (\%)$\uparrow$} \\
            \midrule
            \ffhnet{} \cite{mayer2022ffhnet} &  72.47 &  87.14 &  82.16 & 68.04 \\
            \samplershort{}-PN2 & 82.28 & 98.16  & 81.77 & 81.13 \\
            \samplershort{}-\ac{bps} & \textbf{87.28} & \textbf{98.62} & \textbf{82.64} & \textbf{83.13}\\
            \unidexgrasp{} w/o TTA \cite{Xu_2023_CVPR} & 71.25 & 89.23 & 54.19 & 68.38 \\
            \midrule[0.1pt]
            \ffhnet{}-\ac{esr}-1 &  71.91 & 86.77  &  82.05 & 67.71 \\
            \ffhnet{}-\ac{esr}-2 &  72.16 & 86.86  &   82.21  & 68.29 \\
            \unidexgrasp{} & 72.47 & 85.35 & 61.70 & 67.46\\
            \samplershort{}-\ac{bps}-\ac{esr}-1 & 87.16 & 98.67 & 82.57  & \textbf{83.58} \\
            \samplershort{}-\ac{bps}-\ac{esr}-2 & 87.47  & 98.68 &  82.60 & 83.50\\
            \samplershort{}-\ac{bps}-\ac{egd} & 87.63 & 98.68 &  82.88 & 82.96 \\
            \samplershort{}-\ac{bps}-\ac{egd}-\ac{esr}-1 & 87.41 & 98.68 &  \textbf{82.94}  & 83.08\\
            \samplershort{}-\ac{bps}-\ac{egd}-\ac{esr}-2 & \textbf{87.78} & \textbf{98.77} &  82.90 & 82.96\\
            \bottomrule
        \end{tabular}
    \end{adjustbox}
    \caption{\textbf{Grasp success rate of different methods}. The four first rows exclude grasp refinement, while the bottom eight include it. \samplershort{} represents the \sampler{}. \DexGNDatasetShort{} is short for the DexGraspNet dataset test split. For \unidexgrasp{}, w/o TTA means without the test-time contact-based optimization. $\uparrow$: The higher, the better.}
    % \vspace{-10px}
    \label{tab:sim_exp_esr_egd}
\end{table}

\begin{table}[t]
    \centering
    \begin{adjustbox}{max width=\linewidth}
         \begin{tabular}{lccccccc}
            \toprule
            % \textbf{Method} & \multicolumn{3}{c}{\textbf{Dataset}} \\
            % \textbf{Method} & $H$ mean$\uparrow$ & $H$ std$\downarrow$ \\
            \textbf{Method} & \multicolumn{2}{c}{\textbf{Diversity}} & \multicolumn{4}{c}{\textbf{Max penetration depth (cm)}} \\
            \cmidrule(l){2-7}  & $H$ mean$\uparrow$ & $H$ std$\downarrow$ & \textbf{MultiDex$\downarrow$} & \textbf{EGAD! $\downarrow$} & \textbf{\DexGNDatasetShort{} $\downarrow$} & \textbf{KIT $\downarrow$}\\
            \midrule
            \ffhnet{} &  5.50 &  0.57 & 2.750 & 0.755 & 1.719 & 1.304\\
            \samplershort{}-PN2 & \textbf{5.82} &  \textbf{0.40}  & 2.932 & \textbf{0.750} & 1.557 & 1.145   \\
            \samplershort{}-BPS & 5.79  & 0.42 & \textbf{2.601} & 0.756 & 1.574 & \textbf{1.003} \\
            \unidexgrasp{} & 5.63 & 0.53  & 2.835 & 1.663 & 2.018 & 1.341 \\
             \DexGNSamplerShort{}\cite{dexgraspnet} & 5.77 & 0.40  & 3.519 & 1.604 & \textbf{1.232} & 1.785\\
            \bottomrule
        \end{tabular}
    \end{adjustbox}
    \caption{\textbf{Grasp diversity and max penetration depth}. For diversity, we average the diversity of all datasets. \DexGNSamplerShort{} represents DexGraspNet sampler. $\uparrow$: The higher, the better. $\downarrow$: The lower, the better.}
    % \vspace{-10px}
    \label{tab:diversity}
\end{table}

\begin{figure}[ht]
    \centering
\includegraphics[width=\linewidth]{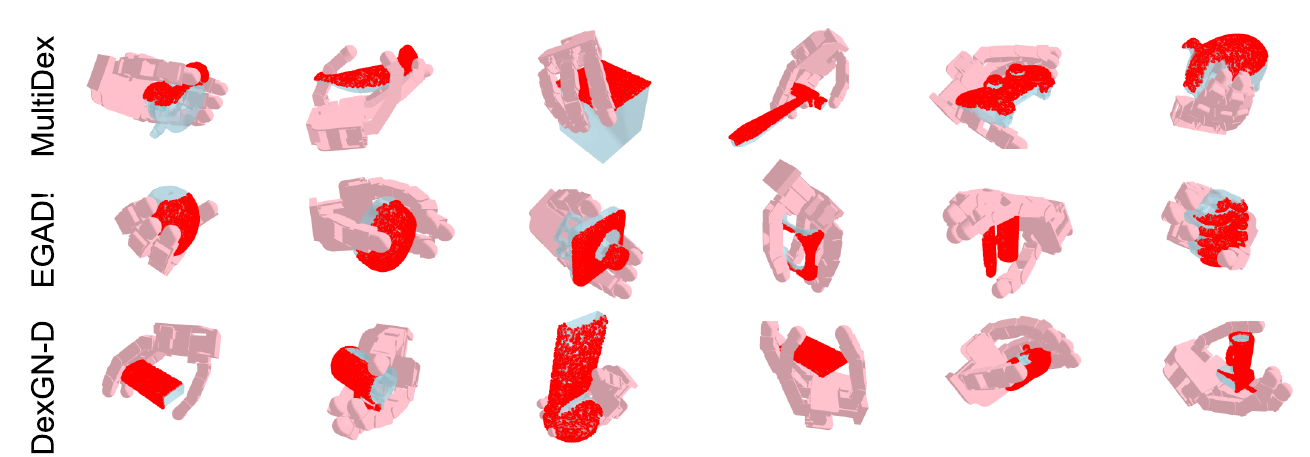}
\caption{ \textbf{Qualitative evaluation of \sampler{}-\ac{bps} on objects from different datasets.} The grasps, shown in pink, are generated on the partially observed point clouds, shown in red, rendered from the complete meshes, shown in blue.}
\label{fig:exp_sim}
\end{figure}

To address the second question, 
%we evaluated both \ffhnet{} and \sampler{} using either the \pointnet{} or \ac{bps} encoding. 
we compared \sampler{} using the \pointnet{} and the \ac{bps} encoding against \ffhnet{} and the grasp proposal network from \unidexgrasp{} \cite{Xu_2023_CVPR}. It is worth pointing out that \unidexgrasp{} requires complete object point clouds, which is not the case for \methodname{} nor \ffhnet{}. The simulation results are presented in \tabref{tab:sim_exp_esr_egd}. These results demonstrate that \sampler{}, irrespective of the encoding method, consistently outperformed the \ffhnet{} generator across all test datasets. The superior performance underscores the diffusion models' effectiveness in synthesizing successful grasps. Moreover, \sampler{}-\ac{bps} performed much better than \sampler{}-PN2, indicating that the \ac{bps} encoding is less affected by point cloud irregularities, including missing parts and non-uniform density distribution. Finally, \figref{fig:exp_sim} shows some grasps generated by the  \sampler{}-\ac{bps} on the three datasets.

Apart from the success rate, we also report the grasp diversity and max penetration depth proposed in \cite{dexgraspnet} in \tabref{tab:diversity}. We excluded the Q1 grasp quality metric from \cite{dexgraspnet} because, as reported in \cite{dexgraspnet}, that metric is unreliable when the max penetration depth exceeds 5mm, which is true for all our methods across all datasets. Following the definition in \cite{dexgraspnet}, the diversity metric is calculated as the average entropy across all joints. We add the diversity and max penetration depth of the training data generated using DexGraspNet as a reference. The results indicate that \sampler{} outperforms both FFHNet and \unidexgrasp{}, highlighting its ability to generate diverse grasps with relatively low penetration depth.

\subsubsection{Grasp Refinement}

To answer question 3, we assessed how grasp refinement affected the grasp success rate. The results, presented in \tabref{tab:sim_exp_esr_egd}, show that \ac{esr} improves grasp success rate for both \methodname{} and \ffhnet{}. For the diffusion-based models, \ac{egd} showed marginal improvements, consistent with findings in diffusion-based robot rearrangement scenarios \cite{liu2023structdiffusion}. The two-stage refinement process \ac{esr}-2 led to slightly better grasp success rates than \ac{esr}-1, giving some evidence to support our hypothesis that it is better to refine global grasp parameters before local ones.

We also evaluated the combination of \ac{egd} and \ac{esr} (\ac{egd}-\ac{esr}) on \methodname{} by first applying \ac{egd} followed by \ac{esr}. Although the combination did achieve the highest grasp success rate on the MultiDex and EGAD! datasets, the improvement is, again, marginal compared to the other methods. We hypothesize the reason refinement does not increase the grasp success rates more is because the average grasp classification accuracy, even for the best model, is only 80.91\%.

\subsection{Real Robot Experiment}

\begin{table*}[ht]
    \centering
    % \vspace{9pt}
    \begin{adjustbox}{max width=\linewidth}
    \setlength{\tabcolsep}{2pt}
         \begin{tabular}{lcccccc}
            \toprule
            \textbf{Real-world experiment}& \multicolumn{6}{c}{\textbf{\# of successful grasps per method and object}} \\
            \cmidrule(l){2-7}
            Object & \ffhnet{} & \ffhnet{}-\ac{esr}-2 & \samplershort{}-\ac{bps} & \samplershort{}-\ac{bps}-\ac{esr}-2& \samplershort{}-\ac{bps}-\ac{egd} & \samplershort{}-\ac{bps}-\ac{egd}-\ac{esr}-2 \\ \midrule
            1. Crackerbox            & 2/5 & 4/5 & 4/5 & 5/5 &  4/5 & 4/5\\
            2. Sugar box              & 3/5 & 4/5 & 4/5 & 4/5 &  4/5 & 5/5\\ 
            3. Mustard bottle         & 3/5 & 3/5 & 4/5 & 5/5 &  3/5 & 3/5\\ 
            4. Bleach cleanser        & 2/5 & 2/5 & 4/5 & 4/5 &  5/5 & 4/5\\ 
            5. Sprayer                & 2/5 & 2/5 & 3/5 & 4/5 &  4/5 & 4/5\\
            6. Metal mug              & 2/5 & 1/5 & 2/5 & 3/5 &  1/5 & 3/5\\
            7. Goblet                 & 3/5 & 3/5 & 5/5 & 5/5 &  5/5 & 4/5\\
            8. Toy plane              & 0/5 & 0/5 & 0/5 & 0/5 &  0/5 & 0/5\\
            9. Pringles               & 2/5 & 3/5 & 2/5 & 1/5 &  3/5 & 2/5\\
            \midrule 
            % \cmidrule(l){2-7}
            Avg. success rate         & 42.22\% & 48.89\% & 62.22\% & \textbf{68.89\%} & 64.44\% & 64.44\%\\
            \midrule 
            Time (ms)                   & \textbf{2} & 490 & 1692 &  2188 & 2548 & 3036 \\ 
            % \midrule 
            
            \bottomrule
        \end{tabular}
    \end{adjustbox}
    \caption{\textbf{Experimental results for real-world grasping.} \samplershort{} represents \sampler{}.}
    \label{tab:real_exp}
    \vspace{-9pt}
\end{table*}

To answer the last question, we evaluated how successful \methodname{} was at picking objects from real noisy object point clouds using real robotic hardware. The hardware used is shown in \figref{fig:pull_figure} and includes an Allegro Hand attached to the end of a Franka Panda robot and a Kinect V3 placed in a third-person perspective to capture $\matr{O}$. We used a motion capture system to do the extrinsic camera calibration.

\begin{figure}[t]
    \centering
\includegraphics[width=0.9\linewidth]{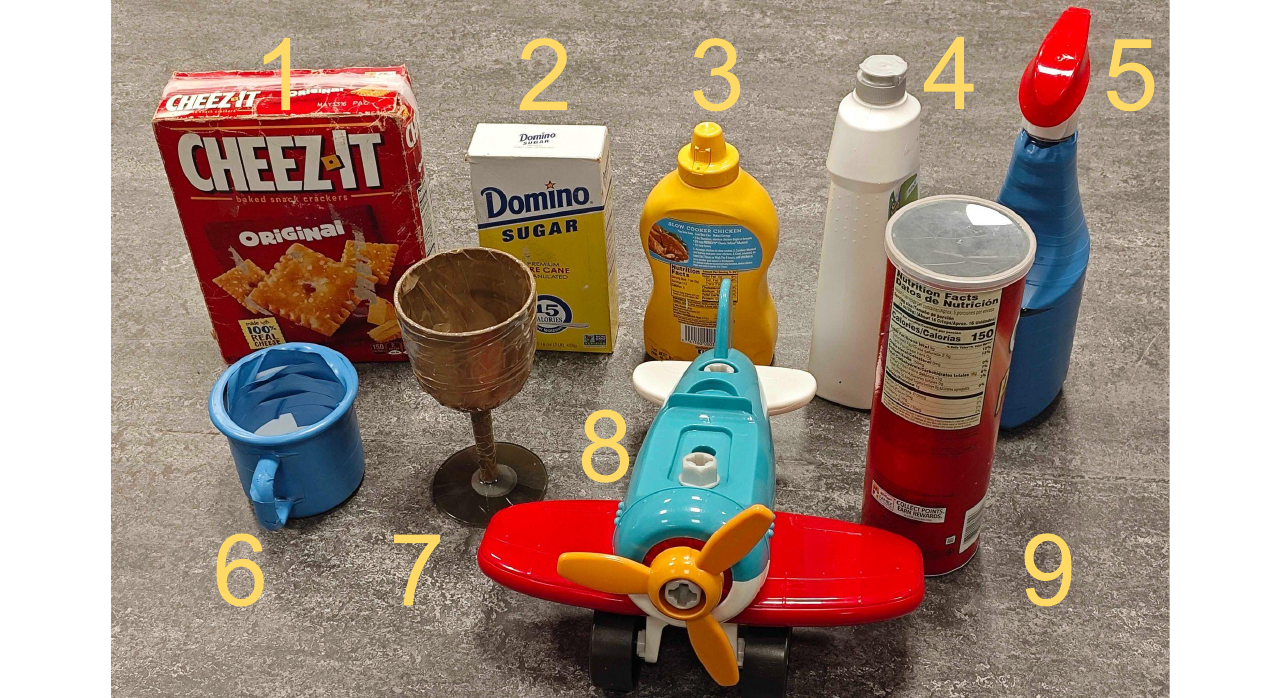}
\caption{\textbf{Experimental objects}. From 1 to 9: cracker box, sugar box, mustard bottle, bleach cleanser, sprayer, metal mug, goblet, toy plane, and pringles.}
\vspace{-10px}
\label{fig:real_obj}
\end{figure}

We chose 9 different test objects, all shown in \figref{fig:real_obj}, based on their variation in shape and size. Each object was placed in the robot's workspace in five distinct orientations: 0\textdegree, 72\textdegree, 144\textdegree, 216\textdegree, and 288\textdegree. All methods generated and ranked 200 grasps per object and orientation. Of these grasps, the best kinematically feasible one was executed on the robot. A grasp was successful if the robot picked the object and kept it grasped while moving to a predefined pose where it rotated the last joint $\pm$90\textdegree.  

\begin{figure}[t]
    \centering
\includegraphics[width=0.9\linewidth]{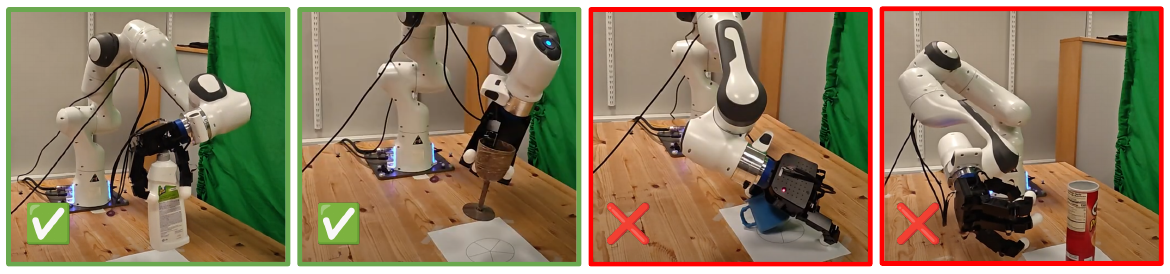}
% \caption{\textbf{Successful and failed grasp examples} from \sampler{}-\ac{bps}-\ac{esr}-2. The first two images successfully grasp the bleach cleanser and the goblet, while the latter two show different failure cases on the metal mug and the pringles.}
\caption{\textbf{Successful and failed grasps from \sampler{}-\ac{bps}-\ac{esr}-2}. The first two images successfully grasp the bleach cleanser and the goblet, while the latter two show different failure cases on the metal mug and the pringles.}
\vspace{-10px}
\label{fig:exp_real_suc_fail}
\end{figure}

We evaluated the following methods: \ffhnet{}, \ffhnet{}-\ac{esr}-2, \methodname{}-\ac{bps}, \methodname{}-\ac{bps}-\ac{esr}-2, \methodname{}-\ac{bps}-\ac{egd} and \methodname{}-\ac{bps}-\ac{egd}-\ac{esr}-2. We omitted \unidexgrasp{} as it requires complete object point-clouds. It is worth highlighting that these methods are only trained on simulated data. The experimental results are summarized in \tabref{tab:real_exp}. Like the simulation results, these results show that \methodname{} consistently outperformed \ffhnet{} across all test objects. The best method for picking real-world objects was \sampler{}-\ac{bps}-\ac{esr}-2, and some of the grasps it produced are shown in \figref{fig:exp_real_suc_fail}.  Interestingly, no methods successfully grasped the toy plane, mainly because the Allegro hand was not strong enough to keep the object secure when the last joint rotated $\pm$90\textdegree.
%, which is demonstrated in website.
%the accompanying video\footnote{\url{https://yulihn.github.io/DexDiffuser_page/}}.

The average grasp success rates in the real world are also consistently lower than in simulation. One rather apparent reason for the decline in performance is that in the real world, grasps are generated on noisy point clouds compared to the clean simulated ones. Another reason is the presence of obstacles that impede the robot's ability to attain the desired target position. For instance, the first failed grasp in \figref{fig:exp_real_suc_fail} is because the gripper collided with the table when closing, while the second failure is because the robot could not reach the intended grasp position due to joint limits. Neither of these failure cases is present in the simulation.

We also report the inference time in \tabref{tab:real_exp}. The iterative denoising process makes the diffusion-based methods slower at inference than \ffhnet{}. For instance, the best performing \methodname{} (DexS-BPS-ESR-2) is 3.5 times slower than the best performing \ffhnet{} (FFHNet-ESR-2).

\section{Conclusion}\label{sec:conclusion}

This paper addressed the problem of dexterous grasping under partial point cloud observations. Our proposed solution, \methodname{}, consists of a conditional diffusion-based grasp sampler and grasp evaluator, \ac{bps}-encoded point clouds, and two different grasp refinement strategies. All experiments on the 16-\ac{dof} Allegro hand demonstrated that \methodname{} outperformed \ffhnet{} with an average of 9.12\% and 19.44\% higher grasp success rate in simulation and real robot experiments, respectively. 

The limitations of \methodname{} are the sim-to-real gap, not accounting for environmental constraints such as collision objects, and the relatively long grasp sampling time. Still, \methodname{} is one of few data-driven dexterous grasping methods that generate grasps directly on partial point clouds and is evaluated on real hardware. Based on the improvements brought by the diffusion model, we believe such methods can be used to solve other challenging dexterous manipulation tasks, including in-hand manipulation, dexterous grasping in clutter, and joint motion and dexterous grasp planning. Addressing these challenges and the limitations of our method highlights promising future research directions.

\balance
\bibliographystyle{IEEEtran}
\bibliography{references}

\end{document}